# Phrasing for UX: Enhancing Information Engagement through Computational Linguistics and Creative Analytics

**Nimrod Dvir**

Department of Information Systems and Business Analytics
State University of New York at Albany
1400 Washington Ave, Albany, NY 12222, USA
ndvir@albany.edu

**Abstract**

This comprehensive study explores the dynamic interplay between textual attributes and Information Engagement (IE) on digital platforms, unveiling the significant role of computational linguistics and creative analytics in optimizing user interaction with online content. Central to our inquiry is the development of the READ model, which quantifies key textual predictors—representativeness, ease of use, affect, and distribution—to forecast user engagement levels. Through methodologically rigorous evaluations, including A/B testing and randomized controlled trials, we validate the model's effectiveness in enhancing key IE dimensions: participation, perception, and perseverance.

Based on a corrected analysis of the test set, the model demonstrates strong predictive performance across the outcomes of participation (accuracy: 0.94, F1-score: 0.884), perception (accuracy: 0.85, F1-score: 0.698), perseverance (accuracy: 0.81, F1-score: 0.603), and overall IE (accuracy: 0.97, F1-score: 0.872). While participation shows excellent results, perception and perseverance have slightly lower recall and F1-scores, indicating some difficulty in accurately identifying all true positive cases across dimensions. Empirical findings indicate that strategic textual modifications, informed by the READ model's insights, lead to substantial improvements in user engagement. Specifically, the study demonstrates that adjustments aiming for higher representativeness and positive affect significantly increase the selection rate by approximately 11%, enhance the evaluation average from 3.98 to 4.46, and boost the retention rate by nearly 11%. These statistically significant enhancements validate the critical influence of linguistic factors on IE. This research contributes both theoretically and practically, providing a framework for assessing and enhancing engagement potential of digital texts. The insights garnered have broad implications across contexts like education, health, and media, equipping content creators with strategies for more engaging information.

## 1    Introduction

Information transitions from raw data to interpreted meaning through transformation processes, evolving into knowledge. This underscores the critical role of information as a precursor to knowledge, rather than knowledge itself (Zins, 2007; Frické, 2009). In digital environments, symbols, letters, words, and phrases have the potential to contribute to knowledge formation, necessitating effective communication and optimal information presentation for Information Systems (IS) success (Delone & McLean, 2003; Venkatesh & Bala, 2008; ISO, 2019). Engagement, defined as the emotional, cognitive, and behavioral connection between users and technological resources, has emerged as a key metric for evaluating user experience (UX), reflecting user interaction depth with a system (O'Brien et al., 2020; Attfield et al., 2011; O'Brien & Cairns, 2016).

The digitization of communication through Information and Communication Technologies (ICT) has revolutionized information conveyance, demanding engaging and effective digital content to ensure successful knowledge transmission and user retention (Beaudry, 2005; Dvir, 2018). Information Engagement (IE) has gained prominence, focusing on the quality of user-system interactions and the impact of digital content design on user decision-making and UX (ISO, 2019; O'Brien, 2020). IE is crucial in enhancing user interactions across domains such as education, government, and industry, aiming to foster meaningful user engagement with digital text (Choi et al., 2018; Feng et al., 2020; Han et al., 2022).

Failure to achieve IE with digital text hinders content producers, yet overcoming this challenge is complicated by a lack of engaging information experience guidelines (Blythe, 2005; Overbeeke et al., 2003). Limited research on IE development has resulted in a scarcity of systematic approaches for its initiation, sustainment, and improvement (O'Brien, 2017; O'Brien & Toms, 2016). Recent advancements in computational linguistics and natural language processing (NLP) have created opportunities to explore systematic, computational, and automatic approaches for creating, evaluating, and improving digital text (Kang, 2020; Dvir & Gafni, 2019).

Traditionally, crafting the right message in digital text has been viewed more as an art than a science. However, this study posits that enhancing IE through strategic word selection can be systematically approached, computationally analyzed, and automated. This paradigm shift towards utilizing creative analytics represents a move towards improving texts based on data-driven insights, offering promising avenues for systematically measuring, predicting, manipulating, and enhancing IE.

Despite technological advancements, there remains a gap in research dedicated to exploring these capabilities for directly augmenting IE systematically and computationally. Existing literature seldom focuses on the informational content—specifically, the effect of phrasing and word choice—on IE and decision-making, nor on how nuanced word selection can substantially boost IE. This gap underscores a critical need for research that investigates both technological and linguistic dimensions influencing user engagement.

By focusing on the impact of phrasing through the lens of word choice, this study aims to fill a significant void in the literature, offering new perspectives on enhancing digital text to foster greater user interaction and engagement. This study seeks to address the gap by examining the effect of phrasing on IE and leveraging computational linguistics and NLP to develop predictive and prescriptive models aimed at optimizing text engagement.

### Objectives

1. To conceptualize and define IE, identifying its key dimensions through an interdisciplinary literature review.
2. To develop the READ (Representativeness, Ease-of-use, Affect, and Distribution) framework as a predictive model for assessing word-level engagement.
3. To create a prescriptive model utilizing NLP for the automatic substitution of more engaging synonyms, enhancing text engagement.

**Research Methodology**

This research is structured around three primary studies:

1. Exploratory Study: Assessing the impact of phrasing on IE through randomized controlled trials.

2. Predictive Model Development: Employing the READ framework to predict engagement levels based on word attributes.

3. Prescriptive Model Implementation: Utilizing NLP and AI to systematically substitute words to enhance IE.

**Significance**

The anticipated findings are expected to significantly contribute to the field of user engagement, highlighting the influence of phrasing on IE and providing a novel, systematic approach for enhancing digital text engagement. By integrating computational linguistics with analytical creativity, this research addresses a gap in literature and offers practical tools for content creators and information system designers to improve digital content quality and engagement. Ultimately, this study aims to transform digital experiences and interactions across various domains by optimizing linguistic choices to maximize user engagement.

## 2 Literature Review

### 2.1 The Imperative of Information Engagement in Information Systems

The influence of user engagement with digital content has been extensively documented across various sectors, highlighting the challenge of capturing user interest amidst diverse motivations (Dvir, 2020; O'Brien, 2020). Engagement is characterized as an immersive experience that necessitates cognitive and psychological investment, significantly shaped by the design of information and the expressiveness of user interfaces (Dvir, 2018; Mollen & Wilson, 2010; O'Brien, 2011). Despite the recognized importance of creating engaging content, there remains a notable scarcity of comprehensive strategies for effectively achieving this goal (O'Brien, 2016).

Information Engagement (IE) represents the depth of user interaction with digital content, incorporating behavioral, cognitive, and emotional aspects (Attfield et al., 2011). It extends the concept of user engagement to encompass meaningful interactions with technology (O'Brien & Toms, 2008), essential for assessing information quality and system efficacy in sectors such as education, healthcare, marketing, and governance (Bardus et al., 2016; Jiang et al., 2016). However, research has primarily concentrated on the theoretical dimensions of IE, providing limited guidance on practical approaches to enhancing user engagement (O'Brien, 2017).

## 2.2    Exploring the Dimensions of Information Engagement

Investigations into IE have unveiled three critical dimensions: participation, perception, and perseverance (Dvir, 2022; O'Brien & Toms, 2008), which elucidate the facets of user engagement with digital content:

**Participation** involves the behavioral component, observable through user actions such as sharing and commenting, as well as passive engagements like reading. This dimension is quantified by metrics like click-through rates and engagement time, revealing nuanced levels of user interaction (Dolan et al., 2016).

**Perception** focuses on attitudinal factors, influenced by users' subjective assessments of content usability, relevance, and aesthetics. Tools like the User Engagement Scale (UES) are employed to evaluate this dimension, underscoring the emotional and cognitive drivers behind engagement (O'Brien et al., 2018).

**Perseverance** reflects the enduring impact of engagement, illustrating how information is retained and applied after interaction. This dimension highlights the depth of cognitive engagement, inferred through content analysis, where higher levels of perseverance indicate deeper and more lasting engagement (Dvir, 2022).

## 2.3    Determinants of Information Engagement

Information Engagement (IE) is influenced by a constellation of factors, including user characteristics, technological attributes, and the inherent qualities of the information presented. This study narrows its focus to textual phrasing—a flexible aspect of content known to significantly impact IE. This choice intersects with the concept of analytical creativity, highlighting the potential for the systematic enhancement of texts through computational linguistics. Despite text's ubiquity in digital interfaces, research into the effects of linguistic nuances on IE is limited. Studies have explored aspects such as readability (Gofman et al., 2009), emotional connotation (Stieglitz & Dang-Xuan, 2012), and semantic associations (Dvir & Gafni, 2018), yet lack a comprehensive framework for integrating these elements.

The gap underscores the underexplored potential of Natural Language Processing (NLP) advancements, which enable detailed textual analysis. Dvir and Gafni (2019) have emphasized that NLP can inform engagement-centric content strategies, though its application remains sparse. Our research aims to bridge this divide by examining textual determinants of IE and developing a systematic approach for content optimization.

### 2.3.1   *Information Engagement and Analytical Creativity*

The landscape of creativity, particularly within Artificial Intelligence (AI) and computational linguistics, is evolving. Our literature review delves into the conceptual foundations of creativity, the rise of analytical creativity, and the implications of computational models for enhancing user engagement through text. Creativity, as defined by Kaufman and Beghetto (2009), involves producing outputs that are both novel and valuable. This broad definition encompasses a spectrum from everyday problem-solving ("little-c creativity") to significant innovations ("Big-C creativity"), further elaborated by the Four C model of creativity which distinguishes between personal insights ("mini-c") and professional expertise ("Pro-c"). These distinctions highlight AI's potential to augment creativity at various levels.

Analytical creativity views creativity as a structured exploration within a defined space (Ding et al., 2024), challenging the traditional perception of creativity as an intangible inspiration and suggesting a methodical approach to understanding and replicating creative processes. It aims to unravel the mechanisms behind creative outputs, blending human intuition with algorithmic precision to scale creativity.

### 2.3.2   *Computational Linguistics and User Engagement*

The intersection of computational linguistics and user engagement presents a promising path for employing analytical creativity. By analyzing and refining textual content, computational linguistics offers a systematic way to enhance the creative allure of digital texts. Although studies have investigated how textual features such as sentiment, complexity, and novelty influence user engagement (Xu et al., 2020), research is scant on the predictive and prescriptive capacities of computational models to systematically improve text's creative quality.

### 2.3.3 Enhancing Information Engagement through Linguistic Features

Our research posits that specific word choices, owing to their cognitive, affective, and semantic properties, have varying engagement potentials. Insights from cognitive psychology suggest that processing fluency and emotional reactions to words significantly influence attention, comprehension, and memory retention (Alter & Oppenheimer, 2009). Computational linguistics enables the quantitative assessment of these features (Narayanan et al., 2013).

Empirical evidence supports that subtle shifts in word choice can lead to significant variations in user behavior and perceptions (Kahneman & Tversky, 1979), underscoring the profound effects of linguistic optimization on IE.

### 2.3.4 Predictive and Prescriptive Models for Creativity Enhancement

The emergence of AI and NLP technologies, including GANs and models like GPT-4, has opened new avenues for mimicking human-like creativity in text generation (Vaswani et al., 2017; Goodfellow et al., 2014). Despite these technologies' potential, there is a crucial research gap in identifying engaging textual features and systematically altering text to elevate creativity and engagement. Our study seeks to fill this void by leveraging creative analytics to predict and enhance digital text engagement, contributing to the emerging domain of analytical creativity in digital contexts.

## 2.4 Theoretical Grounding

This research integrates insights from User Engagement Theory (UET) and Cumulative Prospect Theory (CPT) to investigate user engagement and the nuanced role of cognitive biases in information processing.

### 2.4.1 User Engagement Theory (UET)

UET provides a comprehensive view of user engagement, conceptualized as a cyclical process spanning four phases: point of engagement, sustained engagement, disengagement, and potential re-engagement, each influenced by intrinsic motivations and interactions with technology (O'Brien, 2011). This theory suggests engagement is initiated by aesthetic or novelty appeal and is maintained or potentially re-engaged through continued interaction. The User Engagement Scale (UES), which assesses engagement through metrics such as aesthetic appeal, focused attention, perceived usability, and reward, operationalizes these concepts (O'Brien & Toms, 2008), offering a measurable framework for the experiential aspects of user interaction.

### 2.4.2   *Cumulative Prospect Theory (CPT)*

Originating from behavioral economics, CPT illuminates decision-making under uncertainty, focusing on the effects of framing and cognitive biases, such as representativeness and availability heuristics, on choices (Kahneman & Tversky, 1979). It divides decision-making into framing and valuation phases, underscoring the influence of presentation and perception on outcomes. CPT identifies two cognitive strategies: intuitive, quick but prone to biases, and reasoned, slower but more deliberate (Tversky & Kahneman, 1974), emphasizing the complexity of human judgment.

- **The Framing Effect** demonstrates how presentation changes perceptions and decisions, showing that different framings can lead to diverse outcomes (Tversky & Kahneman, 1981).

- **Heuristics**, such as representativeness (Kahneman & Tversky, 1972), availability (Tversky & Kahneman, 1973), affect (Finucane et al., 2000), and fluency (Alter & Oppenheimer, 2009), serve as mental shortcuts that influence engagement and decision-making. While efficient, these shortcuts can introduce errors but also offer opportunities to enhance engagement by aligning with natural cognitive tendencies.Contrary to rational actor theories, CPT acknowledges cognitive limitations and contextual influences on behavior, providing a comprehensive framework for identifying hidden drivers of user engagement in digital contexts where information overload is common. This theory's consideration of cognitive biases offers invaluable insights for enhancing digital content engagement by leveraging the following:

1. **Representativeness Heuristic**, which impacts preferences for content resembling mental prototypes of relevance or trustworthiness (Kahneman & Tversky, 1974).

2. **Availability Heuristic**, influencing perceptions of relevance or importance based on ease of recall, with implications for engagement shaped by media exposure or recency (Tversky & Kahneman, 1973).

3. **Affect Heuristic**, demonstrating how emotions significantly influence decisions, suggesting that emotionally charged content is more engaging due to its impact on risk and benefit perceptions (Finucane et al., 2000).

4. **Fluency Heuristic**, indicating a preference for easily processed information, suggesting straightforward texts engage users more effectively by reducing cognitive strain (Alter & Oppenheimer, 2008).

These heuristics underpin the 'READ' framework, illustrating the interplay between cognitive biases and user engagement with digital content. By understanding how these biases influence perceptions and behaviors, the framework aims to predict and enhance user engagement through strategic content optimization.

## 2.5    Research Gaps and Research Questions

The advent of computational linguistics and natural language processing (NLP) technologies has opened new avenues for the systematic, computational, and automatic analysis and enhancement of digital text (Dvir, 2019). Despite these technological advancements, there remains a notable deficiency in research focused on measuring, predicting, manipulating, and—crucially—enhancing Information Engagement (IE) in a systematic and computational manner. Further, literature has largely overlooked the impact of textual dimensions, specifically the influence of word choice on IE and decision-making, which represents a significant gap. This study seeks to address these deficiencies by exploring the effect of phrasing on IE, the predictive power of textual features for engaging word selection, and the potential for systematic, computational text modification using computational linguistics, guided by the following research questions:

- **R₁:** How does phrasing impact Information Engagement (IE)?

- **R₂:** Can engaging words be predicted based on their textual features?

- **R₃:** Can text be systematically and computationally modified to enhance IE using computational linguistics?

These questions aim to bridge the identified gaps by leveraging computational linguistics to understand and enhance the engagement potential of digital text, contributing to both theoretical knowledge and practical applications in the field.

### 3    Theoretical Framework and Hypothesis Development

This section delineates the theoretical framework that underpins this study, aiming to coalesce key themes of interest such as critical factors, variables, constructs, and their interrelationships (Miles et al., 2014). The development of this framework was influenced by Webster and Watson's (2002) approach, where an inductive method was employed to generalize and abstract common properties from specific instances, thereby formulating general concepts. This theoretical synthesis, drawing from both domain literature and foundational theories, seeks to comprehensively address the posed research questions.

### 3.1    Influence of Phrasing on Information Engagement (IE)

Leveraging insights from the literature review, this study highlights the pivotal role of linguistic framing in Information Engagement (IE). It posits that the manner in which information is presented, particularly through word choice, is fundamental in shaping user engagement. Integrating User Engagement Theory (UET) with Cumulative Prospect Theory (CPT), we argue that linguistic framing significantly influences IE, suggesting that minor variations in phrasing can substantially affect engagement levels.

Based on the principle that word choice is a crucial determinant of user engagement outcomes, we propose the following hypotheses:

- **H1a:** *Variations in phrasing, particularly in the choice of words for presenting identical information, will significantly affect user participation.*

- **H1b:** *These variations will notably alter user perception.*

- **H1c:** *Furthermore, such variations will influence user perseverance.*

**Operational Definitions:**

- **Participation:** Measured by metrics such as interaction frequency or the propensity for selecting specific words/phrases.

- **Perception:** User assessments of information quality, relevance, and credibility, as influenced by word choice.

- **Perseverance:** The extent to which users maintain engagement, recall, or are influenced by information over time.

## 3.2    Interrelations Among IE Dimensions

Drawing on insights from UET and informed by Information Behavior Theory (IBT), we hypothesize a synergistic relationship among the dimensions of IE, suggesting their interconnectedness:

- **H2a:** *Participation in IE positively correlates with perception.*

- **H2b:** *Participation in IE positively correlates with perseverance.*

- **H2c:** *Perception in IE positively correlates with perseverance.*

These sub-hypotheses aim to clarify the intricate dynamics between IE's dimensions, promoting a holistic understanding of user engagement that includes participation, perception, and perseverance.

Furthermore, these hypotheses acknowledge the nuanced impact of linguistic choices across IE's dimensions. The study leverages synset theory, which posits that the unique cognitive and emotional resonances of synonyms can elicit varying levels of engagement (Miller, 1995).

## 3.3    Developing a Predictive Model: The READ Model

The READ Model marks a significant advancement in predicting Information Engagement (IE) by quantitatively analyzing textual attributes through four dimensions: Representativeness, Ease-of-use, Affect, and Distribution. It utilizes computational linguistics to evaluate the engagement potential of words and phrases, focusing on their functionality, emotion, fluency, familiarity, and findability.

**Phrasing for UX**

Drawing on Cumulative Prospect Theory (Kahneman & Frederick, 2002; Tversky & Kahneman, 1992), we identify four key attributes of information engagement: representativeness, ease of use, affect, and distribution, relating to perceived usefulness, processing fluency, emotionality, and familiarity. These attributes allow for the systematic assessment of linguistic signals that predict engagement.

1. **Representativeness** assesses how closely a new stimulus mirrors an established standard, affecting how individuals categorize and assimilate information (Kahneman & Frederick, 2002). It involves semantic relation analysis to evaluate equivalency, hierarchy, and associative links between words and concepts.

2. **Ease-of-use** prioritizes text that is straightforward and easy to comprehend, influencing decision-making, perception, and memory (Alter & Oppenheimer, 2008; Tversky & Kahneman, 1974). Metrics like the Flesch–Kincaid readability tests assess textual simplicity and cognitive accessibility.

3. **Affect** pertains to the emotional impact of words or phrases, significantly affecting decision-making and engagement levels (Finucane et al., 2000). Sentiment analysis categorizes text to reflect the emotional tone.

4. **Distribution** focuses on the breadth of a word's use, with cognitive biases suggesting a preference for easily retrievable or recognizable information (Kahneman & Frederick, 2002; Tversky & Kahneman, 1992). Word frequency metrics indicate a word's familiarity and overall accessibility.

**Hypotheses Based on the READ Model**

- *H1: Levels of representativeness, ease-of-use, affect, and distribution in words predict their engagement potential, with higher scores correlating with increased user engagement.*

- *H2: Among synonyms conveying identical information, those scoring higher in representativeness, ease-of-use, affect, and distribution will be more engaging.*

**Summary Table of the READ Model Attributes**

| Attribute | Definition | Factor | Measurement |
|---|---|---|---|
| Representativeness | Degree of similarity to a standard | Familiarity | Semantic relation |
| Ease-of-Use | Complexity and cognitive load | Fluency | Simplicity |
| Affect | Emotional association | Feeling | Sentiment analysis |
| Distribution | Frequency and recognizability | Availability | Saliency/significance |

The READ model is presented as a comprehensive framework for evaluating and predicting the engagement potential of textual content, emphasizing the operationalization of its components and the development of hypotheses to test within the study's context.

**3.4    Prescriptive Model of Information Engagement - Application of the READ Model**

The culmination of this research is the application of the READ Model in a user-centered, data-driven approach to identify and leverage significantly engaging words—terms with a high potential to enhance user engagement. This process employs Text Data Mining (TDM), computational linguistics, and Natural Language Processing (NLP) to evaluate the engagement potential of words and phrases across the dimensions of representativeness, ease-of-use, affect, and distribution.

Our approach involves a thorough analysis of word impact on IE's key dimensions: participation, perception, and perseverance. The model explores the potential to boost user engagement by replacing less engaging synonyms with more engaging alternatives, maintaining the core message intact. This strategy validates the hypothesis that specific word choices significantly influence user engagement levels. For instance, modifying a title from "Is the Pirate Party the new maven of media accountability or a self-serving movement?" to "Is the Pirate Party the new star of media accountability or a self-serving movement?" demonstrates how nuanced phrasing adjustments can markedly improve IE without changing the content's intended message.

### Framework and Strategy

The conceptual framework introduced herein provides a holistic strategy for understanding and enhancing IE at the intersection of creative analytics and computational linguistics. By conceptualizing IE as a multifaceted construct and employing both predictive and prescriptive models, our research lays the groundwork for innovative digital content optimization methods aimed at augmenting user engagement. This advanced methodology underscores the critical role of language and word choice in influencing digital interactions, offering practical guidance for content creators and information system designers to maximize engagement.

### Hypotheses Development

The formulation of H3 and H4 is bolstered by an integration of cognitive psychology, computational linguistics techniques, and empirical insights into user engagement. This comprehensive approach not only highlights the importance of textual optimization in boosting IE but also anticipates a consistent effect across various engagement dimensions, providing a solid theoretical and empirical basis for these hypotheses.

- **H3:** *Optimizing textual phrasing by substituting less engaging terms with more engaging alternatives, as identified by predictive models, enhances overall information engagement.*

- **H4:** *The impact of linguistic optimization on information engagement is consistent across the dimensions of participation, perception, and perseverance.*

A tangible example of these hypotheses in action is the revision of the aforementioned title to enhance IE, illustrating the substantial influence of subtle phrasing changes on user engagement without modifying the content's original intent.

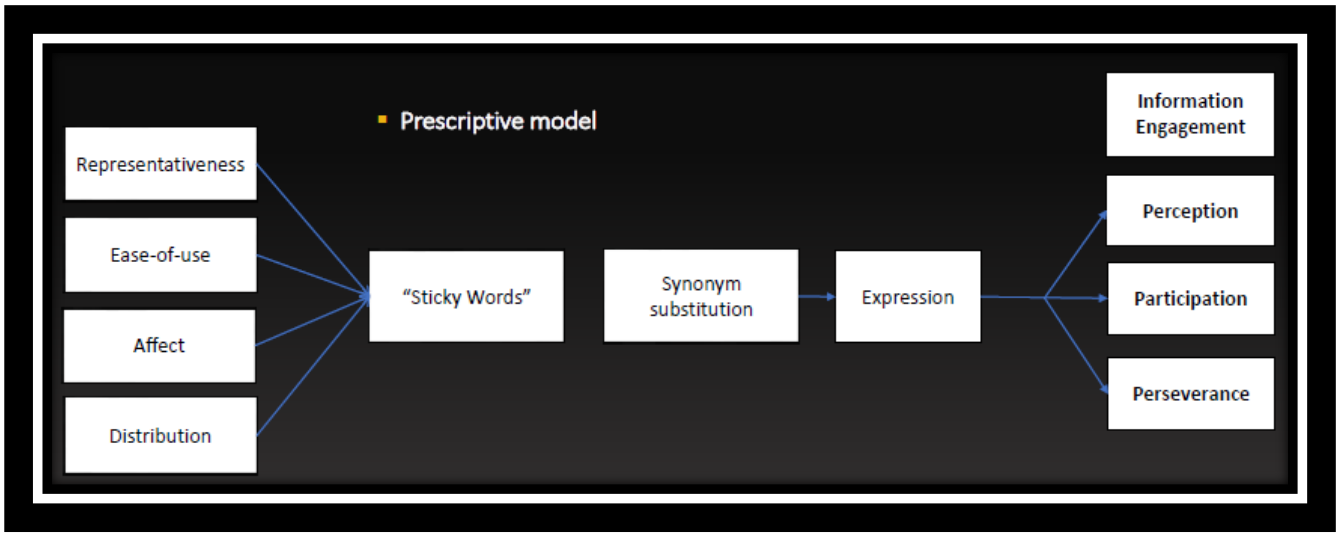

# 4    Study 1: Exploratory Analysis on Phrasing Impact

## 4.1    Study Design

This study investigates the relationship between phrasing variations and the dimensions of Information Engagement (IE) - participation, perception, and perseverance. It examines how different wordings (independent variable) influence IE (dependent variable).

## 4.2    Methodology

### 4.2.1    Instruments

A selection of 250 synonym sets (synsets) from WordNet was utilized to explore the nuances of linguistic variation on user engagement. These synsets were chosen for their balance of relatedness and distinctiveness, aiming to uncover subtle differences in engagement elicited by varied word choices

Here is an example of a few of the word pairs that were randomly chosen:

| Word₁ | Word₂ | Synset | Definition |
|---|---|---|---|
| abused | maltrated | abused.a.02 | Subjected to cruel treatment |
| star | maven | ace.n.03 | Someone who is dazzlingly skilled in any field |
| quick | nimble | agile.s.01 | Moving quickly and lightly |
| rich | plenteous | ample.s.02 | Affording an abundant supply |
| annoying | nettlesome | annoying.s.01 | Causing irritation or annoyance |
| art | prowess | art.n.03 | A superior skill learned by study, practice, and observation |
| gone | deceased | asleep.s.03 | Dead |
| zombie | automaton | automaton.n.01 | Someone who acts or responds in a mechanical or apathetic way |

| Word₁ | Word₂ | Synset | Definition |
|---|---|---|---|
| greedy | avaricious | avaricious.s.01 | Immoderately desirous of acquiring something, typically wealth |
| king | magnate | baron.n.03 | A very wealthy or powerful businessman |
| mother | engender | beget.v.01 | Make children |
| bubbling | belching | burp.v.01 | Expel gas from the stomach |
| fighter | belligerent | combatant.n.01 | Someone who fights or is fighting |
| computerization | cybernation | computerization. n.01 | The control of processes by computer |
| cut | shortened | cut.s.03 | With parts removed |
| lady | gentlewoman | dame.n.02 | A woman of refinement |
| death | demise | death.n.04 | The time at which life begins to end and continuing until death |

within a controlled lexical framework. QualtricsXM, a comprehensive cloud-based survey platform, facilitated the survey administration and data collection. This platform ensured the collection of participant demographics, device usage, and survey responses, maintaining data integrity by allowing only one completion per participant.

### 4.2.2 Procedure and Measurements

Participants engaged with an online survey presenting a randomized selection of words from the chosen synsets, ensuring varied and randomized exposure across the dataset.

**Perception Measurement -** Perception was evaluated using statements adapted from the User Engagement Scale (UES), focusing on sensory appeal, attention, usability, and reward. Participants rated their agreement on a 5-point Likert scale, with negative items reverse-coded for analysis consistency.

*Table 1: Perception Evaluation Statements (Adapted from UES)*

| Code | Statement |
|---|---|
| EA | This word appealed to my senses. |
| EA-n | This word is not engaging. |
| FA | This word drew my attention. |
| FA-n | I wasn't focused while reading this word. |
| PU | This word was easy to understand. |
| PU-n | This word was difficult to understand. |
| RW | Reading this word was rewarding. |
| RW-n | Reading this word was not worthwhile. |

Responses ranged from 1 (strongly disagree) to 5 (strongly agree), applying reverse coding to negative statements.

**Participation Measurement -** Engagement rates were derived from binary selection responses, capturing participants' willingness to engage with specific words.

**Perseverance Measurement -** Recall of previously shown words was the metric for perseverance. Responses were coded as remembered (1) or not (0), employing ChatGPT-4 for advanced matching against the original list, accommodating spelling variations and providing refined retention insights.

### 4.2.3   Participants

Participants were undergraduate students from a large research university in the United States, with recruitment and survey methodologies approved by the University at Albany Institutional Review Board (IRB Study No. 22X113) for ethical compliance.

### 4.2.4   Sampling and Randomization

Qualtrics software verified unique completions, with random presentation of each dataset word to participants, controlling for participant characteristics. This design aimed to minimize biases, enhancing the study's reliability in assessing linguistic impacts on user engagement. This methodology combines meticulous selection with innovative measurement techniques, aiming to deepen the understanding of word choice's influence on user interaction with text.

## 4.3   Findings

### 4.3.1   Exploratory Data Analysis

**Phrasing for UX**

In this exploratory analysis, 8,050 users participated, each exposed to 10 words (5 pairs), yielding 80,500 observations with each word presented 161 times. The demographic breakdown showed 41.2% females and 58.8% males, with an average age of 22.1 years (SD = 1.388). Most participants (75.3%) were aged 17–22, 79.0% were native English speakers, and device usage comprised 70.1% laptops, 15.6% mobile devices, and 14.3% desktops.

Randomization's effectiveness was verified through statistical tests to ensure comparable demographic distributions across word samples, crucial for isolating the impact of phrasing on engagement. Chi-square tests showed no significant differences in gender distribution across word samples ($\chi^2(1, N = 8,050) = 2.56$, p = .11), indicating successful demographic balancing. Chi-square tests also found no significant differences in device usage across samples ($\chi^2(2, N = 8,050) = 5.42$, p = .07), ensuring uniform distribution across devices. One-way ANOVA confirmed no significant age differences among groups (F(249, 8,050) = 1.02, p = .42), validating the randomization's effectiveness.

These analyses affirm that the observed engagement variations are attributable to the phrasing rather than demographic differences, enhancing the findings' generalizability.

**H1a: Variations in Phrasing and User Participation**

Analysis of participation rates (average rate of 21.8) using a 2x2 chi-square test across 250-word pairs revealed significant differences in 29.2% of pairs. This result supports H1a, demonstrating that phrasing variations significantly impact user participation.

**H1b: Variations in Phrasing and User Perception**

Cronbach's alpha validated the survey instrument's reliability for assessing perception (.85), with an average perception score of 2.44 (SD = 0.84). A Z-test differentiating high (score ≥4.0, achieved by 12.3% of observations) from low perception rates identified significant perception score differences in 32% of word pairs, affirming that phrasing variations significantly influence user perception.

**H1c: Variations in Phrasing and Perseverance**

**Phrasing for UX**

Perseverance, or recall rate, was 8% across 80,500 observations. A chi-square test comparing recall successes and failures across word pairs found significant differences in 28% of pairs, indicating that phrasing significantly affects word recall, supporting the hypothesis on perseverance impact.

**Hypothesis 2: Interrelations Among IE Dimensions**

Utilizing insights from User Engagement Theory (UET) and Information Behavior Theory (IBT), this study posited a synergistic relationship among the dimensions of Information Engagement (IE): participation, perception, and perseverance. These dimensions were hypothesized to be interconnected and mutually reinforcing, reflecting a comprehensive understanding of user engagement.

- **H2a:** A significant positive correlation was observed between participation in IE and perception ($r(1) = .805$, $p < .05$), affirming that increased participation is associated with enhanced perception. This result supports H2a, highlighting a direct relationship between these dimensions of engagement.

- **H2b:** The study also found a positive correlation between participation in IE and perseverance ($r(1) = .666$, $p < .05$), supporting H2b. This indicates that higher participation levels correlate with greater perseverance among users.

- **H2c:** Furthermore, a positive correlation between perception in IE and perseverance was established ($r(1) = .661$, $p < .05$), corroborating H2c. This suggests that improvements in perception can lead to increased perseverance in engagement.

**Chi-squared Analysis**

A series of Chi-squared tests further examined the interdependencies within these engagement dimensions. **Participation and Perception:** A significant association was found ($\chi^2(1, N = 250) = 51.70$, $p < .001$), confirming the interconnectedness of these dimensions. **Participation and Perseverance:** Results indicated a significant relationship ($\chi^2(1, N = 250) = 43.13$, $p < .001$), underscoring the link between active engagement and long-term retention. **Perception and Perseverance:** A significant correlation was also observed ($\chi^2(1, N = 250) = 41.79$, $p < .001$), highlighting the role of perception in fostering enduring engagement.

**Phrasing for UX**

Despite significant interrelations among some dimensions, not all word pairs showed uniform significance across all dimensions: 60.8% (152 pairs) showed no significant differences across the three dimensions. 20.0% (50 pairs) exhibited significant differences in all dimensions of IE. The remaining pairs showed varied significance, with some combinations of dimensions being significant and others not.

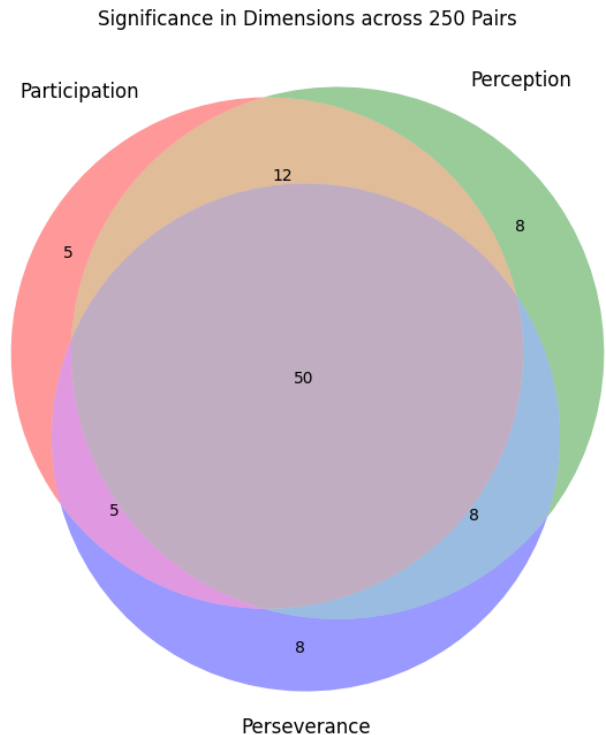

### 4.3.2   Discussion

These findings affirm the hypothesized synergistic relationships among the dimensions of IE, suggesting that participation, perception, and perseverance are significantly associated and mutually reinforcing. The varied significance across different dimensions and word pairs underscores the complexity of user engagement and the nuanced impact of linguistic choices. This study's insights into the interrelations among IE dimensions provide a robust framework for understanding and enhancing user engagement through strategic content optimization.

These findings illustrate the significant influence of subtle linguistic variations on all dimensions of Information Engagement (IE): participation, perception, and perseverance. The data underscore the importance of word choice in digital content strategies to enhance user engagement, providing empirical support for the READ model's predictive capacity and the effectiveness of linguistic optimization in content creation.

## 5    Study 2 - Predictive Model Development

Following the foundational insights from Study 1, which established that specific word choices could significantly influence Information Engagement (IE) across participation, perception, and perseverance dimensions, this phase aims to pinpoint textual predictors for enhancing IE effectively. Our approach involves a novel framework that systematically identifies engaging information and integrates a predictive model, leveraging computational linguistics to measure and predict IE attributes precisely.

The core objective of Study 2 was to develop and validate the READ model, positing that the textual attributes of representativeness, ease of use, affect, and distribution are key predictors of user engagement with digital content.

### 5.1    Measurements

In developing the predictive model, the process began with feature extraction, transforming textual information into a numerical format suitable for Natural Language Processing (NLP) analysis. The bespoke READ program, leveraging Python and various NLP libraries such as the Natural Language Toolkit (NLTK), was instrumental in quantifying the attributes of words according to the READ model's dimensions: Representativeness, Ease-of-use, Affect, and Distribution.

#### 5.1.1    *Representativeness Measures:*

1. **Definitions (Senses):** Utilizing WordNet, this measure quantifies the multiplicity of meanings a word possesses (polysemy) by counting its distinct senses, reflecting the breadth of a word's semantic field. For example, the word "star" is associated with 12 synsets in WordNet, demonstrating a higher polysemy compared to "maven," which links to only one synset.

2. **Hypernyms:** Identifies broader terms that encompass more specific words, like "color" for "red," providing insight into the hierarchical structure of language.

3. **Hyponyms:** Specifies terms that are more detailed instances of a broader category, further illustrating the word's positioning within a semantic hierarchy.

### 5.1.2 Ease-of-use Measures:

1. **Length:** The number of characters in a word, calculated using NLTK, indicating potential complexity.

2. **Syllable Count:** Determines complexity by counting syllables within a word, affecting readability.

3. **Flesch Reading Ease Score:** A readability formula that scores text based on its simplicity, correlating higher scores with easier comprehension.

### 5.1.3 Affect Measures:

1. **Sentiment Score:** Utilizing SentiWordNet 3.0, this metric assesses the emotional tone a word conveys, ranging from positive to negative, and quantifies its emotional impact.

### 5.1.4 Distribution Measures:

1. **Frequency:** A measure of how commonly a word appears across various sources, providing a comprehensive frequency score.

2. **Zipf Frequency:** Applies the base-10 logarithm of a word's occurrences per billion words to evaluate its commonality.

### 5.1.5 Data Splitting for Model Training and Testing

The dataset was divided into training and test subsets, adhering to an 80/20 split. This allocation involved using 200 words for training the model, while the remaining 50 words served as the test set. This split was strategic, ensuring that the model could be trained on a substantial portion of the data before being validated against an unseen subset to assess its predictive accuracy. After developing the READ model, the next phase involved testing its predictive capabilities on the dataset collected from Study 1.

**Phrasing for UX**

The aim was to evaluate the probability of a word being significantly higher in participation, perception, perseverance, and overall Information Engagement (IE), capturing those that excel in all dimensions.

During the training phase, logistic regression models were fitted using the 200-word training set. Each model utilized the quantified attributes of words—such as their sentiment scores, readability levels, frequency measures, and semantic richness—as predictors. The objective was to establish a statistical relationship between these attributes and the words' engagement scores across the IE dimensions.

Upon training, the model was then applied to the 50-word test set to predict their engagement potential. The model's predictions were compared against actual engagement outcomes to evaluate its accuracy. This step was crucial in determining the model's effectiveness in identifying words with a high potential for boosting user engagement.

## 5.2 Findings

### 5.2.1 Participation predictors

Here's a concise summary of the regression results for predicting significant participation in Information Engagement (IE), formatted into a table for clarity and space efficiency:

| Variable | B | SE B | β | t | p |
|----------|---|------|---|---|---|
| (Constant) | .207 | .001 | - | 386.743 | .000 |
| Hypernyms | .001 | .000 | .292 | 51.891 | .000 |
| Hyponyms | -4.169E-6 | .000 | -.017 | -5.073 | .000 |
| Definitions Synsets | .000 | .000 | -.233 | -39.534 | .000 |
| EmotionalityMax2 | .028 | .001 | .513 | 47.049 | .000 |
| EmotionalitySum | -.021 | .000 | -.473 | -43.413 | .000 |
| Length (len) | -.001 | .000 | -.112 | -13.750 | .000 |
| Flesch Reading Ease | 2.634 E-6 | .000 | .013 | 1.719 | .086 |
| Syllables (sylla) | -.002 | .000 | -.115 | -13.593 | .000 |
| WnZipf | .005 | .000 | .419 | 85.397 | .000 |

- **Hypernyms** and **WnZipf** show positive coefficients, suggesting that broader categorizations and commonality (Zipf's frequency) significantly enhance participation.

- **Hyponyms, Definitions Synsets,** and **Length** negatively influence participation, indicating that specificity, polysemy, and word length may detract from user engagement.

- **EmotionalityMax** exhibits a strong positive effect, while **EmotionalitySum** shows a significant negative impact, highlighting the complex role of emotional content.

- **Syllables** also have a negative association with participation, suggesting that simpler words (fewer syllables) are more engaging.

- **Flesch Reading Ease** shows a positive but not statistically significant (p = .086) relationship with participation, indicating a marginal influence of readability on engagement.

These results highlight the multifaceted impact of textual attributes on user engagement, validating the predictive power of the READ model for participation outcomes. The significant

predictors—Hypernyms, WnZipf, and EmotionalityMax—offer actionable insights for optimizing digital content to enhance user participation effectively.

### 5.2.2 Perception predictors

Here's a concise summary of the regression results for predicting significant perception significance in Information Engagement (IE), formatted into a compact table for clarity and space efficiency:

| Variable | B | SE B | β | t | p |
|---|---|---|---|---|---|
| (Constant) | 2.297 | 0.005 | - | 443.730 | 0.000 |
| Definitions Synsets | 0.000 | 0.000 | -0.003 | -0.704 | 0.481 |
| Hypernyms | -0.005 | 0.000 | -0.185 | -40.455 | 0.000 |
| Hyponyms | 0.000 | 0.000 | 0.034 | 12.669 | 0.000 |
| PosMax | 0.126 | 0.005 | 0.161 | 23.911 | 0.000 |
| EmotionalityMax | 0.047 | 0.006 | 0.073 | 8.084 | 0.000 |
| NegMax | -0.113 | 0.005 | -0.143 | -24.326 | 0.000 |
| Length (len) | -0.019 | 0.001 | -0.241 | -36.046 | 0.000 |
| Flesch Reading Ease | -0.001 | 0.000 | -0.328 | -51.952 | 0.000 |
| Syllables (sylla) | -0.041 | 0.001 | -0.223 | -32.292 | 0.000 |
| WnZipf | 0.108 | 0.001 | 0.722 | 177.654 | 0.000 |
| WnFreq | -175.563 | 1.842 | -0.302 | -95.297 | 0.000 |

**Key Insights:**

- **Hypernyms, NegMax, Length,** and **Flesch Reading Ease** show significant negative associations with participation, suggesting that broader terms, negative sentiments, longer words, and overly simple texts may detract from engagement.

- **Hyponyms, PosMax, EmotionalityMax,** and **WnZipf** are positively linked to participation, indicating that specificity, positive emotions, emotional resonance, and commonality enhance user engagement.

- **WnFreq**'s negative coefficient highlights a complex relationship between word frequency and engagement, with rare or unique words potentially reducing participation.

- The significant **p** values across most variables affirm their impact on participation, underscoring the multifaceted influences of textual attributes on user engagement.

This table efficiently encapsulates the statistical analysis, illustrating the nuanced effects of various textual features on Information Engagement, as predicted by the READ model.

### 5.3 IE – significant higher on all three dimensions

Here's a concise summary of the logistic regression results for predicting significant overall Information Engagement (IE), which encompasses being significant across all dimensions (participation, perception, and perseverance):

| Feature | Coefficient | P-value |
|---|---|---|
| const | -2.0611 | 0.000 |
| DefinitionsSynsets | -0.0375 | 0.101 |
| Hypernyms | -0.1023 | 0.000 |
| Hyponyms | 0.0184 | 0.164 |
| PosMax | 0.1312 | 0.000 |
| NegMax | -0.0686 | 0.000 |
| Syllables | -0.0700 | 0.025 |
| Length | -0.0410 | 0.239 |
| Frequency | -0.2195 | 0.000 |
| wnzipf | 0.5569 | 0.000 |

- **Hypernyms, NegMax, Syllables,** and **Frequency** have significant negative coefficients, indicating that broader categorizations, negative sentiments, higher syllable count, and lower frequency are associated with reduced overall IE.

- **PosMax** and **wnzipf** show positive coefficients, suggesting that positive sentiment and higher Zipf frequency values (commonality) significantly enhance overall IE.

- **DefinitionsSynsets** and **Hyponyms** coefficients are not statistically significant ($p > 0.05$), indicating that the multiplicity of meanings and specificity may not have a clear impact on overall IE within this model.

- **Length** also shows a negative coefficient but is not statistically significant ($p = 0.239$), suggesting that while there might be a trend towards shorter words enhancing IE, this result is not conclusive.

These results provide valuable insights into the textual predictors of Information Engagement, emphasizing the importance of positivity, commonality, and simplicity in enhancing engagement across all dimensions. The significant predictors identified through this logistic regression analysis can inform content optimization strategies for maximizing user engagement.

## 5.4    Model's Predictive Performance Overview

The model's prediction accuracy was assessed to be 92.5% for both Logistic Regression and Random Forest models, indicating a strong predictive performance.

Based on the corrected analysis of the test set, here are the performance metrics for each of the outcomes (Participation, Perception, Perseverance, and IE) with a threshold of 0.5 for classifying the probability of being significantly higher on all dimensions:

| Category | Accuracy | Precision | Recall | F1-Score |
|----------|----------|-----------|--------|----------|
| Participation | 0.94 | 1.00 | 0.793 | 0.884 |
| Perception | 0.85 | 0.846 | 0.595 | 0.698 |
| Perseverance | 0.81 | 0.688 | 0.537 | 0.603 |
| IE | 0.97 | 0.895 | 0.85 | 0.872 |

These metrics suggest the following about the model's performance on the test set:

- **Participation**: The model shows excellent performance in predicting significant one-sided modifications in participation, with high accuracy, precision, recall, and F1-score.

- **Perception**: The model performs well in predicting significant one-sided modifications in perception, though with slightly lower recall and F1-score compared to participation.

- **Perseverance**: The model's performance in predicting significant one-sided modifications in perseverance shows the lowest recall and F1-score among the dimensions, indicating a challenge in accurately identifying true positive cases.

- **IE**: The model demonstrates strong performance in predicting cases where all three dimensions are significantly higher, with high accuracy and an F1-score over 80%.

| Category | Observed Positives (TP + FN) | Predicted Positives (TP + FP) | True Positives (TP) | False Positives (FP) | False Negatives (FN) |
|---|---|---|---|---|---|
| Participation | 29 | 23 | 23 | 0 | 6 |
| Perception | 37 | 26 | 22 | 4 | 15 |
| Perseverance | 41 | 32 | 22 | 10 | 19 |
| IE | 20 | 19 | 17 | 2 | 3 |

These results indicate that the model is quite effective in distinguishing between word pairs with significantly higher rates of participation, perception, and perseverance, as well as identifying cases where all three dimensions are significantly higher. The high precision across all outcomes suggests that the model's positive predictions are reliable. However, the recall values, especially for perception and perseverance, suggest room for improvement in correctly identifying all actual positive cases.

The model showcases strong overall predictive capabilities, with particularly high precision, suggesting that when it predicts an increase in engagement, those predictions are highly reliable. However, variations in recall across different dimensions suggest that further refinement may be needed to ensure all positive cases are accurately captured, particularly in perception and perseverance dimensions.

This detailed performance assessment underscores the model's effectiveness in distinguishing word pairs that significantly enhance engagement, providing a solid foundation for further optimization and application in content strategy and development.

## 6    Study 3: Prescriptive Model Testing

Study 3 presents a thorough empirical validation of the READ model, assessing its impact on elevating Information Engagement (IE) through strategic textual modifications. We modified titles taken from the New York Times based on synanon substitution using the OPENAI API and analytics from the

**Phrasing for UX**

READ model. We made sure not to change the meaning of the title. Below is an example of a few of the titles:

| Original Title | Modified Title |
|---|---|
| Half of **Palestinians in Gaza** Are at **Jeopardy** of **Famine**, **United Nations** Cautions | Half of **Gazans** Are at **Risk** of **Starving**, U.N. Warns |
| How to **Begin** the New Year? Satisfy the **Ocean Deity**. | How to **Start** the New Year? Keep the **Sea Goddess** Happy. |
| What's **Draining** Your Vigor? | What's **Sapping** Your Energy? |
| Day 1: A 5-Minute **Technique** for Increased Vigor | Day 1: A 5-Minute **Trick** for More Energy |
| Engage in our energy **assessment**. | Take our energy **quiz**. |
| The Acquisition of Language | The Learning of Language |
| Digital Metamorphosis in Intricate Systems | Digital Transformation in Complex Systems |
| Ascertain the research imperatives for emergency care within the Western Cape province of South Africa: A unanimity study | Determining the research priorities for emergency care within the Western Cape province of South Africa: A consensus study |
| An Examination of Monotonous Speech Patterns in Audiobooks | An Examination of Repetitive Speech Patterns in Audiobooks |
| A Longitudinal Study of Lackluster Student Performance in STEM | A Repeated Measures Study of Poor Student Work in STEM |
| The Neuroscience of Language | The Brain Science of Language |
| The Consequences of Physical Exertion on Cognitive Function | The Impact of Physical Activity on Mental Function |

| The Essence of Existence | The Meaning of Life |
|---|---|
| The Contribution of Genetics in Mental Agility | The Role of Genetics in Intelligence |
| The Consequences of Social Isolation on Psychological Well- being | The Effects of Social Isolation on Mental Health |
| The Association Between Stress and Ailment | The Relationship Between Stress and Disease |

## 6.1    Objectives

1. **:** Utilize the READ model predictions for automatic textual modifications.

2. **Evaluate Impact on IE Dimensions:** Measure the effect of textual modifications on perception, participation, and perseverance.

3. **Ascertain Model's Effectiveness:** Compare engagement levels between original and modified texts.

## 6.2    Methodology

- **Design:** Employed A/B testing (randomized controlled trials) to compare engagement between original and modified texts.

- **Participants:** Engaged undergraduate students from a large U.S. research university, with IRB approval and informed consent obtained.

- **Measures:** Direct (surveys and feedback) and indirect (behavioral analytics) methods were used to assess engagement levels, same as in study 1.

## 6.3    Key Findings

The study demonstrated significant improvements in IE metrics for texts modified using the READ model:

| Metric | T | P | Mean (Original) | SD (Original) | Mean (Modified) | SD (Modified) | P(Sig) |
|---|---|---|---|---|---|---|---|

| | Selection Rate (%) | 30.046 | 2.10e-23 | 47.80 | 2.73 | 58.97 | 2.36 | *** |
|---|---|---|---|---|---|---|---|---|
| | Evaluation Average (out of 5) | 17.948 | 3.02e-17 | 3.98 | 0.21 | 4.46 | 0.19 | *** |
| | Retention Rate (%) | 35.132 | 2.54e-25 | 58.77 | 2.18 | 69.37 | 2.09 | *** |

- **Statistical Significance:** "***" indicates a p-value < 0.001, showing highly significant statistical differences between original and modified conditions across all metrics.

- **Chi-Square Analysis:** Revealed significant associations between sentence modification and both selection and retention rates, indicating that modified sentences were chosen and retained at significantly higher rates than original sentences.

## 6.4   Statistical Tests Summary

- **Paired Samples T-Test:** Significant difference in scores for selection rate, evaluation average, and retention rate between original and modified sentences (p < .001 for all).

- **Chi-Square Test:** Demonstrated a significant association between sentence modification and improvements in selection and retention rates.

## 6.5   Evaluation Success Rates

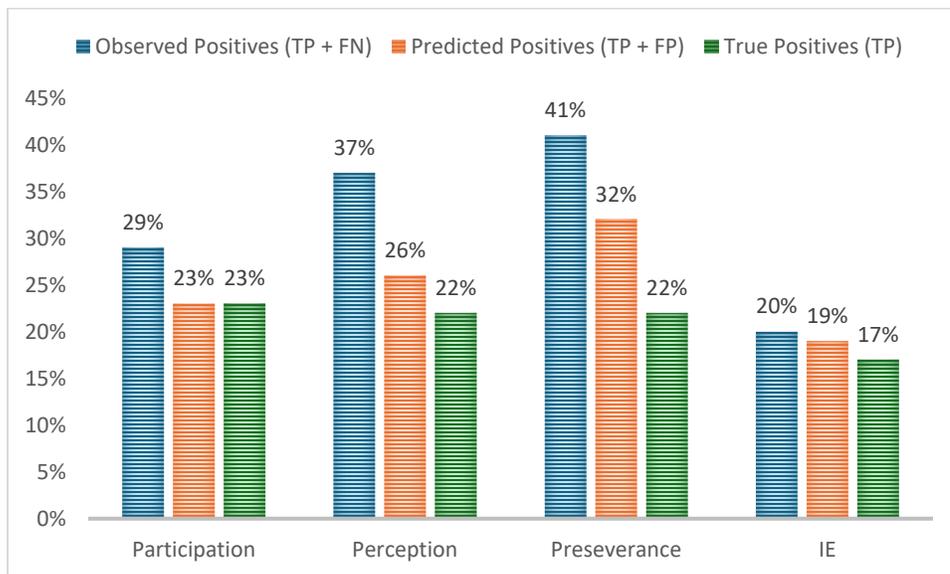

- **Evaluation Success:** 24/30

- **Retention Success:** 21/30

- **Selection Success:** 18/30

## 6.6    Conclusion

Study 3's comprehensive analysis validates the DIMA model's efficacy in significantly improving IE through textual modifications. The findings from both paired t-tests and Chi-square tests confirm that the modifications positively impact all measured dimensions, indicating a robust and statistically significant enhancement in user engagement with digital content.

## 7    Discussion

This research spans three studies to explore how textual attributes influence Information Engagement (IE) using the READ model for prediction and optimization. Study 1 establishes the model's foundation, identifying key textual predictors. Study 2 validates its predictive accuracy, demonstrating a 92.5% success rate in enhancing participation, perception, and perseverance. Study 3 employs A/B testing, confirming that strategic synonym substitutions significantly improve IE metrics across digital content. Collectively, these studies offer a groundbreaking framework for digital content optimization, contributing to both theoretical insights and practical applications in computational linguistics and creative analytics.

## 7.1    Interpretation of Findings

The comprehensive analysis across our studies sheds light on the intricate relationship between textual attributes and Information Engagement (IE), facilitated by the development and empirical testing of the READ model. This research highlights the pivotal role of linguistic factors—representativeness, ease of use, affect, and distribution—in engaging users with digital content, reinforcing the notion that computational linguistics can substantially predict and boost user engagement in digital settings.

Our findings reveal the nuanced impact of word choice on user engagement, demonstrating that while all dimensions of IE (evaluation, selection, and retention) are affected by textual attributes, the degree and nature of this influence vary. Notably, the evaluation dimension exhibited the most pronounced effect, underscoring the importance of how words are perceived in terms of appeal and usability in driving overall information engagement.

Statistically significant variations in IE scores across different words further validate the model's efficacy, indicating clear distinctions in selection, retention, and evaluation scores based on the linguistic composition of the content.

## 7.2     Theoretical Implications

This research enhances the theoretical landscape of IE by providing a detailed exploration of its dimensions and determinants through the lens of computational linguistics. The READ model operationalizes IE, offering a quantitative framework to evaluate engagement potential grounded in textual attributes. Additionally, the application of this model for prescriptive text enhancement introduces a groundbreaking methodological avenue for creative analytics in the realm of digital content optimization.

## 7.3     Practical Implications

Practically, this research equips content creators and digital strategists with robust tools for enhancing communication effectiveness. The READ model, supported by our empirical findings, acts as a catalyst for systematically improving digital text engagement. This model promises to revolutionize content management systems, enabling scalable optimization across various digital content types, from marketing materials to educational resources.

## 7.4     Contributions to Creative Analytics and Computational Linguistics

By marrying creative analytics with computational linguistics, our study pioneers an interdisciplinary approach to enhancing digital user experiences. It exemplifies how analytical creativity, underpinned by computational methods, can dissect and reassemble text to systematically elevate user engagement. This innovative synergy not only advances the fields of information systems and computational linguistics but also opens new pathways for the optimization of digital content, setting a precedent for future research and application in this burgeoning domain.

Phrasing for UX

In sum, the READ model's theoretical and practical contributions underscore its value in navigating the complexities of digital engagement, offering a new paradigm for understanding and improving how users interact with text online. Through this work, we have laid a foundational stone for future explorations into optimizing digital experiences, providing a robust framework for academics and practitioners alike to build upon.

## 7.5    Study Limitations

While our research on the READ model provides significant insights into enhancing Information Engagement (IE) through textual modifications, it is not without limitations. Acknowledging these limitations is crucial for framing the context of our findings and for guiding future research in this domain.

1. **Scope of Textual Attributes:** Our research primarily focused on representativeness, ease of use, affect, and distribution as key textual attributes influencing IE. While comprehensive, this scope may not encapsulate all possible factors that affect engagement, such as context-specific nuances, cultural connotations, or evolving language trends.

2. **Generalizability Across Content Types:** The studies predominantly analyzed article titles, which, although pivotal for engagement, represent only a fraction of digital content. The effectiveness of the READ model across diverse content types, such as long-form articles, social media posts, or interactive content, remains to be fully explored.

3. **User Demographics and Personalization:** The research largely treated the audience as a monolithic entity, without delving deeply into the impacts of demographic factors or individual preferences on IE. Personalization and demographic-specific engagement strategies were not a focus but could significantly influence the model's applicability and effectiveness.

4. **Language and Cultural Considerations:** The study was conducted within the framework of English language content. Language-specific idiosyncrasies and cultural nuances play a critical role in engagement, suggesting that the model's adaptability to different languages and cultures warrants further investigation.

**7.6    Avenues for Future Research**

Given these limitations, several avenues for future research emerge, promising to extend the utility and applicability of the READ model and related frameworks:

1. **Expanding Textual Attribute Analysis:** Future studies could incorporate additional linguistic and non-linguistic attributes into the model, such as narrative style, thematic elements, or visual-textual interplay, to provide a more holistic understanding of IE.

2. **Cross-Content Type Validation:** Research should aim to test and validate the READ model across various digital content formats. This includes examining how textual modifications influence engagement in blogs, videos, podcasts, and interactive media.

3. **Personalization and Demographic Analysis:** Investigating how personalization and demographic factors, including age, culture, and language proficiency, affect IE could offer insights into tailoring content more effectively. This line of research would enhance the model's precision in engaging diverse audiences.

4. **Multilingual and Cross-Cultural Adaptation:** Exploring the model's adaptability and effectiveness in different linguistic and cultural contexts is essential. Future research should aim to adapt and test the READ model across languages and cultures, assessing how global audiences engage with modified content.

5. **Integration with Emerging Technologies:** With the advancement of AI and machine learning, integrating the READ model with these technologies could automate and refine content optimization processes. Research could explore how AI-driven content modification influences user engagement in real-time.

6. **Longitudinal Impact Studies:** Understanding the long-term effects of textual modifications on user engagement and behavior would provide valuable insights into the sustainability of engagement strategies derived from the READ model.

**Phrasing for UX**

By addressing these limitations and exploring these avenues, future research can build on the foundational work presented here, further advancing our understanding of digital engagement and opening new pathways for optimizing online content to meet diverse user needs and preferences.

## 8    Conclusion

Our exploration into the dynamics of Information Engagement (IE) through the lens of the READ model has traversed a significant academic and practical terrain, revealing the profound impact of textual attributes on user interaction with digital content. This journey, encapsulated in a series of methodologically rigorous studies, has not only validated the predictive power of the READ model but also demonstrated its practical efficacy in enhancing digital engagement through strategic textual modifications.

The READ model, grounded in the principles of computational linguistics and creative analytics, has emerged as a pivotal tool for understanding and optimizing IE. By focusing on representativeness, ease of use, affect, and distribution as core textual attributes, our research has illuminated the nuanced ways in which word choice can significantly influence user participation, perception, and perseverance. The empirical validation of the model through A/B testing and randomized controlled trials underscores the tangible benefits of applying the READ model to digital content, with notable improvements in engagement metrics across varied content types.

Theoretically, this research contributes to the expanding body of knowledge on user engagement, offering a nuanced understanding of how textual attributes interact to influence IE. It advances the discourse on computational linguistics and creative analytics, providing a robust framework for future investigations into digital content optimization. Practically, the READ model offers content creators, digital strategists, and information system designers an innovative approach to systematically enhance the engagement quality of texts, from article titles to comprehensive digital narratives.

As we look ahead, the potential for further refining the READ model and exploring its applicability across diverse content types, languages, and cultural contexts is vast. The integration of user demographic factors, personalization strategies, and advanced machine learning technologies presents exciting opportunities for making digital content more engaging and impactful. Continued interdisciplinary research, combining insights from computational linguistics, information systems, and user experience design, will be crucial in navigating the complexities of digital engagement in an increasingly interconnected world.

In conclusion, the journey through the development, testing, and application of the READ model has been both enlightening and invigorating. It highlights the critical role of language in shaping digital experiences and opens new avenues for enhancing user engagement with online content. As we move forward, the insights gleaned from this research will undoubtedly serve as a cornerstone for future endeavors aimed at optimizing digital content, ultimately enriching the user experience in the digital landscape.

**Phrasing for UX**